\documentstyle[twocolumn]{coling}
\title{\bf REFERENCE RESOLUTION USING SEMANTIC PATTERNS IN JAPANESE
NEWSPAPER ARTICLES}
\author{Takahiro Wakao\\
\\
University of Sheffield, Department of Computer Science\\
Regent Court, 211 Portobello St, Sheffield S1 4DP, UK\\
Email: {\em t.wakao@dcs.shef.ac.uk}\\
}
\date{}
\begin{document}
\maketitle
\thispagestyle{empty}
\section{INTRODUCTION}
\indent
Reference resolution is one of the important tasks in natural
language processing.  In Japanese newspaper articles, pronouns are not
often used as referential expressions for company names, but
shortened company names and {\em dousha} (``the same company'') are used
more often (Muraki {\em et al}. 1993).
Although there have been studies of
reference resolution for various noun phrases in Japanese (Shibata {\em et al}.
1990; Kitani 1994), except Kitani's work, they do not clearly show how to
find the referents in
computationally plausible ways for a large amount of data, such as a
newspaper database.  In this paper\footnote[1]{This paper was written
when the author was at the Computing Research Laboratory of New Mexico
State University.  The author has been at University of Sheffield
since January 1994.}, we determine the referents of {\em dousha} and
their locations by hand, and then propose one simple and two
heuristic methods which use semantic information in text such as
company names and their patterns, so as to test these three methods on how
accurately they find the correct referents.\\
\\
\indent
{\em Dousha} is found with several particles such as ``{\em ha}'',
``{\em ga}'', ``{\em no}'', and ``{\em to}'' in newspaper articles.
Those which co-occur with {\em ha} and {\em ga} are chosen for the data
since they are the two most frequent particles when {\em dousha} is in the
subject position in a sentence.  Typically, {\em ha} marks the topic of
the sentence and {\em ga} marks the subject of the sentence.
A typical use of  {\em dousha} is as follows:
\begin{quotation}
\noindent
Nihon Kentakii Furaido Chikin ha,\\
Japan Kentucky Fried Chicken  ha,\\
\\
sekai saidai no piza chien,\\
world's largest pizza chain store,\\
\\
Piza Hatto to teikei wo musubi,\\
Pizza Hut to tie-up establish,\\
\\
kotoshi gogatsu kara zenkoku de\\
starting May this year, nation-wide,\\
\\
takuhai piza chien no tenkai wo \\
pizza delivery chain store  extension\\
\\
hajimesu to happyou shita.\\
begin      announced.\\
\\
sarani {\em dousha ha} furaido chikin no\\
Moreover, the same company fried chicken of\\
\\
takuhai saabisu nimo noridasu.\\
delivery service as well  will start.
\end{quotation}
\indent
A rough translation is:\\
``Kentucky Fried Chicken Japan announced that it had
established a tie-up with the world largest pizza chain store, Pizza
Hut, and began to expand pizza delivery chain stores nation-wide
starting in May this year.  Moreover, {\em the company} will start
delivery of fried chicken as well.''\\
\\
\indent
{\em Dousha} in the second sentence refers to Kentucky Fried
Chicken Japan as  ``{\em the company}'' does in the English translation.  As
shown in this example, some articles contain more than one possible
referent or company, and the reference resolution of {\em dousha}
should identify the referent correctly.
\section{LOCATIONS AND CONTEXTS OF THE REFERENTS}
\indent
Most of the Japanese newspaper articles examined in this study are in the
domain of Joint-Ventures. The sources of the newspaper articles are mostly
{\em the Nikkei} and {\em the Ashahi}. The total number of the
articles is 1375, and there are 42 cases of {\em dousha} with {\em ga}
and 66 cases of {\em dousha} with {\em ha} in the entire set of articles.\\
\\
\indent
The following tables,
{\bf Table} 1 and {\bf Table} 2, show the locations and contexts where the
referents of both subsets of {\em dousha} appear.\\
\newpage
\onecolumn
\begin{center}
{\bf Table} 1  Locations and contexts of the referents of {\em dousha} with
{\em ga}\\
\vspace{4 mm}
\begin{tabular}{|l|l|r|}	\hline
\multicolumn{3}{|c|}{{\em dousha} with {\em ga}} \\ \hline
location	& context	& number of cases \\ \hline
\multicolumn{2}{|c}{Within the same sentence} &
 \multicolumn{1}{|r|}{19} \\ \hline
Subject     & company name $+$ {\em ha} &  7 \\
            & part of the subject $\ast $  &  1 \\
Non-subject & company name $+$ {\em niyoruto} &  3 \\
            & others $\ast \ast \ast$ & 8 \\ \hline
\multicolumn{2}{|c}{In the previous sentence} &
 \multicolumn{1}{|r|}{13} \\ \hline
Subject     & company name $+$ {\em ha} &  8 \\
            & company name $+$ {\em ga} &  1 \\
            & emphasis structure $\ast \ast $ &  1 \\
            & part of the subject $\ast $ &  1 \\
Non-subject & company name $+$ {\em to} &  2 \\ \hline
\multicolumn{2}{|c}{In two sentences before} &
 \multicolumn{1}{|r|}{6} \\ \hline
Subject     & company name $+$ {\em ha} &  5 \\
            & company name $+$ {\em ga} &  1 \\ \hline
\multicolumn{2}{|c}{In previous paragraph} &
 \multicolumn{1}{|r|}{1} \\ \hline
Topic of the paragraph & company name $+$ {\em ha} &  1 \\ \hline
\multicolumn{2}{|c}{In two paragraphs before} &
 \multicolumn{1}{|r|}{3} \\ \hline
Topic of the paragraph & company name $+$ {\em ha} &  3 \\ \hline
\end{tabular}\\
\vspace{6 mm}
{\bf Table} 2  Locations and contexts of the referents of {\em dousha} with
{\em ha}\\
\vspace{4 mm}
\begin{tabular}{|l|l|r|}	\hline
\multicolumn{3}{|c|}{{\em dousha} with {\em ha}} \\ \hline
location	& context	& number of cases \\ \hline
\multicolumn{2}{|c}{Within the same sentence} &
 \multicolumn{1}{|r|}{2} \\ \hline
Subject     & company name $+$ {\em ga} &  1 \\
            & company name $+$ {\em deha} &  1 \\ \hline
\multicolumn{2}{|c}{In the previous sentence} &
 \multicolumn{1}{|r|}{32} \\ \hline
Subject     & company name $+$ {\em ha} &  21 \\
            & emphasis structure $\ast \ast $ &  5 \\
            & part of the subject $\ast $ &  4 \\
Non-subject & others &  2 \\ \hline
\multicolumn{2}{|c}{In two sentences before} &
 \multicolumn{1}{|r|}{17} \\ \hline
Subject     & company name $+$ {\em ha} &  16 \\
            & part of the subject $\ast $ &  1 \\ \hline
\multicolumn{2}{|c}{In three sentences before (in the same paragraph)} &
 \multicolumn{1}{|r|}{2} \\ \hline
Subject     & company name $+$ {\em ha} &  2 \\ \hline
\multicolumn{2}{|c}{In previous paragraph} &
 \multicolumn{1}{|r|}{7} \\ \hline
Topic of the paragraph & company name $+$ {\em ha} &  6 \\
Topic of the paragraph & company name $+$ {\em ga} &  1 \\ \hline
\multicolumn{2}{|c}{In two paragraphs before} &
 \multicolumn{1}{|r|}{2} \\ \hline
Topic of the paragraph & company name $+$ {\em ha} &  2 \\ \hline
\multicolumn{2}{|c}{In three paragraphs before} &
 \multicolumn{1}{|r|}{2} \\ \hline
Topic of the paragraph & company name $+$ {\em ha} &  2 \\ \hline
\end{tabular}
\vspace{6 mm}
\\
Note for Table 1 and Table 2 \\
\vspace{3 mm}
\begin{tabular}{|l|l|}	\hline
$\ast $ & company name referred to is a part of a larger subject noun
phrase. \\ \hline
$\ast \ast$ & company name referred to comes at the end of the \\
 & sentence, a way of emphasising the company name in Japanese. \\ \hline
$\ast \ast \ast$ & company name with {\em to} (with), {\em kara}
(from),\\
 & {\em wo tsuuji} (through), {\em tono aidade} (between or among). \\
\hline
\end{tabular}
\end{center}
\vspace{2 mm}
\twocolumn
\indent
For {\em dousha} with {\em ga} ({\bf Table} 1), the referred company names,
or the referents appear in non-subject positions from time to time,
especially if the referent appears in the same sentence as {\em
dousha} does.  For {\em dousha} with {\em ha} ({\bf Table} 2), compared with
{\bf Table} 1, very few referents are located in the same sentence, and most
of the referents are in the subject position.  For both occurrences of {\em
dousha}, a considerable number of the referents appear two or more
sentences before, and a few of them show up even two or three
paragraphs before.

\section{THREE HEURISTIC METHODS TESTED}
\subsection{Three Heuristic Methods}
\indent
One simple and two heuristic methods to find the referents of {\em dousha}
are described below.  The first, the simple method, is to take
the closest company name, (the one which appears most recently before
{\em dousha}), as its referent
({\bf Simple Closest Method} or {\bf SCM}).  It is used in this paper
to indicate the baseline performance for reference resolution of {\em
dousha}.\\
\\
\indent
The second method is a modified Simple Closest
Method for {\em dousha} with {\em ga}.  It is basically the same
as SCM except that:
\begin{itemize}
\item{if there is one or more company name in the same
sentence before the {\em dousha}, take the closest company name as
the referent.}
\item{if there is a company name immediately followed by {\em ha},
{\em ga}, {\em deha}, or {\em niyoruto} somewhere before {\em dousha},
use the closest such company name as the referent.}
\item{if the previous sentence ends with a company name, thus
putting an emphasis on the company name, make
it the referent.}
\item{if there is a pattern ``company name {\em no} human name
title...'' (equivalent to ``title human name of company name...'' in
English) in the previous sentence, then use the company name as the
referent.
Typical titles are {\em shachou} (president) and {\em kaichou}
(Chairman of Board).}
\end{itemize}
\noindent
The third heuristic method is used for {\em dousha} with {\em
ha} cases.  It is also based on SCM except the following points:
\begin{itemize}
\item{if there is a company name immediately followed by {\em ha},
{\em ga}, {\em deha}, or {\em niyoruto} somewhere before {\em dousha},
use the closest such company name as the referent.}
\item{if the previous sentence ends with a company name, thus
putting an emphasis on the company name, make
it the referent.}
\item{if there is a pattern ``company name {\em no} human name
title...'' (equivalent to ``title human name of company name...'' in
English) in the previous sentence, then use the company name as the
referent.}
\end{itemize}
\indent
The third method is in fact a set of the second method, and both of them
use semantic information (i.e. company name, human name, title), syntactic
patterns (i.e. where a company
name, a human name, or a title appears in a sentence) and
several specific lexical items which come immediately
after the company names.

\subsection{Test Results}
\indent
The three methods have been tested on the development data
from which the methods were produced and on the set of
unseen test data.

\subsubsection{Against the development data}
As mentioned in section two, there are 42 cases of {\em dousha} with
{\em ga} and 66 cases of {\em dousha} with {\em ha}.\\
\\
\indent
For the {\em dousha} with {\em ga} cases, the Simple Closest Method
identifies the referents {\bf 67}\% correctly (27 correct out of 42), and
the second method does so {\bf 90}\% (38 out of 42) correctly.  SCM
misses a number of referents which appear in
previous sentences, and most of those which appear two or more sentences
previously.\\
\\
\indent
For the cases of {\em dousha} with {\em ha}, SCM
identifies the referents correctly only {\bf 52}\% (34 correct out of 66),
however, the third heuristic method correctly identifies {\bf
94}\% (62 out of 66).
\subsubsection{Against the test data}
The test data was taken from Japanese newspaper articles on micro-electronics.
There are 1078 articles, and 51 cases of {\em dousha} with
{\em ga} and 250 cases of {\em dousha} with {\em ha}.  The test has been
conducted against the all {\em ga} cases (51 of them) and the first
100 {\em ha} cases.\\
\\
\indent
For the {\em dousha} with {\em ga} cases, the Simple Closest Method
identifies the referents {\bf 80}\% correctly (41 correct out of 51), and
the second method does so {\bf 96}\% (49 out of 51) correctly.\\
\\
\indent
For the cases of {\em dousha} with {\em ha}, SCM
identifies the referents correctly only {\bf 83}\% (83 correct out of 100),
however, the third heuristic method correctly identifies {\bf
96}\% (96 out of 100).\\
\\
The following table, {\bf Table} 3, shows the summary of the test
results.\\
\begin{center}
{\bf Table} 3   Summary of Test Results\\
\vspace{4 mm}
\begin{tabular}{|l|r|r|}	\hline
 & Development Data & Test Data \\ \hline
\multicolumn{3}{|c|}{{\em dousha} with {\em ga}} \\ \hline
SCM & 67 \% & 80 \% \\ \hline
2nd method & 90 \% & 96 \% \\ \hline
\multicolumn{3}{|c|}{{\em dousha} with {\em ha}} \\ \hline
SCM & 52 \% & 83 \% \\ \hline
3rd method & 94 \% & 96 \% \\ \hline
\end{tabular}
\end{center}
\section{DISCUSSION}
\indent
The second and third heuristic methods show high accuracy in
finding the referents of {\em dousha} with {\em ga} and {\em ha}.
This means that partial semantic parsing (in which key semantic
information such as company name, human name, and title is marked) is
sufficient for reference resolution of important referential
expressions such as {\em dousha} in Japanese.  Moreover, since
the two modified methods are simple, they will be easily implemented by
computationally inexpensive finite-state pattern matchers (Hobbs {\em et al}.
1992; Cowie {\em et al}. 1993).  Therefore, they will be suitable for large
scale text processing (Jacobs 1992; Chinchor {\em et al}. 1993).\\
\\
\indent
One important point to realize is that the second and third
methods, although they are simple to implement,
achieve something that is rather complicated and may be
computationally expensive otherwise.  For example, in order to find
the correct referent of
a given {\em dousha}, you may have to skip one entire paragraph and
find the referent two paragraphs before, or you may have to choose
the right company name from several possible company names which
appear before the given {\em dousha}.  The modified methods
do this correctly most of the time without worrying about
constructing sometimes complicated syntactic structures of the
sentences in the search window for the possible referent.\\
\\
\indent
Another important point is that the modified methods make good use
of post-nominal particles, especially {\em ha} and {\em ga}.
For example, if the referent is located two sentences or
more before, then the referent (the company name) comes with {\em ha}
almost all the time (35 out of 38 such cases for both {\em dousha}).
It seems that if the referent of the {\em dousha} in consideration is
more than a certain distance before, two sentences in this case,
then the referent is marked with {\em ha} most of the time.
Kitani also uses this {\em ha} or {\em ga} marked company names
as key information in his reference resolution algorithm for
{\em dousha} (Kitani 1994).

\section {CONCLUSION}
\indent
The locations and contexts of the referents of {\em dousha} in
Japanese Joint-Venture articles are determined by hand.  Three
heuristic methods are proposed and tested.  The methods which use
semantic information in the text and its patterns show high accuracy
in finding the referents (96\% for {\em dousha} with {\em ga} and
96\% for {\em dousha} with {\em ha} for the unseen test data).
The high success rates suggest that
a semantic pattern-matching approach is not only a valid method but
also an efficient method for reference resolution in the newspaper article
domains.  Since the Japanese language is highly case-inflected, case
(particle)
information is used effectively in these methods for reference
resolution.  How much one can do with semantic pattern matching for
reference resolution of similar expressions such as ``the company'' or
``the Japanese company'' in English newspaper articles is a
topic for future research.

\section{ACKNOWLEDGEMENT}
\indent
I would like to thank the Tipster project group at the CRL for their
inspiration and suggestions.  I would also like to thank Dr. Yorick
Wilks, Dr. John Barnden, Mr. Steve Helmreich, and Dr. Jim Cowie for
their productive comments.
The newspaper articles used in this study are from the Tipster
Information Extraction project provided by ARPA.

\section{REFERENCES}

Chinchor, N., L. Hirschman, and D. Lewis (1993).  Evaluating Message
Understanding Systems: An Analysis of the Third Message Understanding
Conference (MUC-3). {\em Computational Linguistics, 19(3)},
{\em pp.} 409-449.\\
\\
Cowie, J., T. Wakao, L. Guthrie, W. Jin, J. Pustejovsky, and
S. Waterman (1993).  The {\em Diderot} Information Extraction
System.  In the proceedings of {\em The First Conference of the
Pacific Association for Computational Linguistics (PACLING 93)}
Simon Fraser University, Vancouver, B.C. Canada, {\em pp.} 23-32. \\
\\
Jacobs, P.S. (1992).  Introduction: Text Power and Intelligent
Systems.  {\em In} P.S. Jacobs {\em Ed}.,
{\em Text-Based Intelligent Systems}.
Lawrence Erlbaum Associates, Hillsdale New Jersey, {\em pp.} 1-8.\\
\\
Hobbs, J., D. Appelt, M. Tyson, J. Bear, and D. Israel (1992).
SRI International Description of the FASTUS System used for MUC-4.
In the proceedings of {\em Fourth Message Understanding Conference
(MUC-4)}, Morgan Kaufmann Publishers, San Mateo, {\em pp.} 269-275.\\
\\
Kitani, T. (1994).  Merging Information by Discourse Processing
for Information Extraction. In the proceedings of {\em the tenth
IEEE Conference on Artificial Intelligence for Applications},
{\em pp.} 168-173.\\
\\
\\
Muraki, K., S. Doi, and S. Ando (1993).  Context Analysis in
Information Extraction System based on Keywords and Text Structure.
In the proceedings of {\em the 47th National Conference of
Information Processing Society of Japan}, {\em 3-81}. (In Japanese).\\
\\
Shibata, M., O. Tanaka, and J. Fukumoto (1990). Anaphora in
Newspaper Editorials.  In the proceedings of {\em the 40th National
Conference of Information Processing Society of Japan}, {\em 5F-4}. (In
Japanese).

\end{document}